\documentclass[article]{jss}

\usepackage{orcidlink,thumbpdf,lmodern}

\usepackage[utf8]{inputenc}
\usepackage{graphicx}
\usepackage{amsmath}
\usepackage{amsthm}
\usepackage{fancyvrb}
\usepackage{array}
\usepackage{multirow}
\usepackage{multicol}
\usepackage{microtype}
\usepackage{psfrag}
\usepackage{booktabs}
\usepackage{amssymb}
\usepackage{natbib}
\usepackage[verbose]{newunicodechar}
\usepackage{a4wide}
\usepackage{mathtools}
\usepackage{tikz}
\usepackage{nccmath}
\usepackage{float}
\usepackage{microtype}
\usepackage{fontenc}
\usepackage{dsfont}
\usepackage{setspace}
\usepackage[justification=centering]{caption}
\usepackage{roboto}
\usepackage{etoolbox}
\usepackage{acronym}
\usepackage{dirtytalk}
\usepackage{listings} 
\usepackage{xcolor}
\usepackage[english]{babel}
\usepackage{pifont}
\usepackage{makecell}
\usepackage{relsize}
\usepackage{tikz}
\usepackage{todonotes}
\usepackage{breakurl}

\newcommand{\class}[1]{`\code{#1}'}

\newcommand\V[1]  { \mathbf{#1} }
\newcommand\B[1]  { \boldsymbol{#1} }

\newcommand\set[1] {\mathcal{#1}}

\newcommand{\up}[1]{\mathrm{#1}}

\newcommand{\xmark}{\ding{55}}%
\newcommand{\sset}[1]{\mathsmaller{\mathcal{#1}}}
\newcommand\ptr {\mathrm{p}_{\text{tr}}}
\newcommand\pte {\mathrm{p}_{\text{te}}}

\newcolumntype{P}[1]{>{\centering\arraybackslash}p{#1}}
\DisableLigatures[<,>]{encoding=T1,family=tt*}

\acrodef{RFF}[RFF]{random Fourier features}
\acrodef{RRM}[RRM]{robust risk minimization}
\acrodef{ERM}[ERM]{empirical risk minimization}
\acrodef{MRC}[MRC]{minimax risk classifier}
\acrodef{RFE}[RFE]{recursive feature elimination}
\acrodef{AMRC}[AMRC]{adaptive minimax risk classifier}
\acrodef{SVM}[SVM]{support vector machine}
\acrodef{LP}[LP]{linear program}
\acrodef{LR}[LR]{logistic regression}
\acrodef{SOA}[SOA]{state-of-the-art}
\acrodef{DT}[DT]{decision tree}
\acrodef{NN}[NN]{neural network}
\acrodef{SGD}[SGD]{stochastic gradient descent}
\acrodef{OOP}[OOP]{object-oriented approach}
\acrodef{DW-GCS}[DW-GCS]{double-weighting general covariate shift}
\acrodef{KMM}{kernel mean matching}
\acrodef{DW-KMM}{double-weighting \ac{KMM}}

\newcommand{\jose}[1]{{\color{black}#1}}
\newcommand{\kar}[1]{{\color{black}#1}}

\vspace{-0.5cm}
\author{Kartheek Bondugula\\Basque Center for \\Applied Mathematics
   \And Ver\'{o}nica \'{A}lvarez\\ Basque Center for \\Applied Mathematics \And Jos\'{e} I. Segovia-Mart\'{i}n\\ Basque Center for \\Applied Mathematics \AND Santiago Mazuelas \\Basque Center for \\Applied Mathematics, \\IKERBASQUE-Foundation for Science \And Aritz P\'{e}rez \\Basque Center for \\Applied Mathematics}
\Plainauthor{Kartheek Bondugula, Ver\'{o}nica \'{A}lvarez, Jos\'{e} I. Segovia-Mart\'{i}n, Santiago Mazuelas, Aritz P\'{e}rez}

\title{MRCpy: A Library for Minimax Risk Classifiers}
\Plaintitle{MRCpy: A Library for Minimax Risk Classifiers}
\Shorttitle{MRCpy: A Library for Minimax Risk Classifiers}

\Abstract{
Libraries for supervised classification have enabled the wide-spread usage of machine learning methods. Existing libraries, such as \pkg{scikit-learn}, \pkg{caret}, and \pkg{mlpack}, implement techniques based on the classical \ac{ERM} approach. We present a \proglang{Python} library, \pkg{MRCpy}, that implements \acp{MRC} based on the \ac{RRM} approach. The library offers multiple variants of \acp{MRC} that can provide performance guarantees, enable efficient learning in high dimensions, and adapt to distribution shifts. \pkg{MRCpy} follows an object-oriented approach and adheres to the standards of popular \proglang{Python} libraries, such as \pkg{scikit-learn}, facilitating readability and easy usage together with a seamless integration with other libraries. The source code is available under the GPL-3.0 license at \url{https://github.com/MachineLearningBCAM/MRCpy}.
}

\Plainkeywords{supervised classification, robust risk minimization, high dimensions, concept drift, covariate shift, minimax risk classifiers, Python}

\begin{document}
\sloppy
\shortcites{montiel2021river}
\shortcites{gomes2017adaptive}

\section{Introduction} \label{sec:intro}

\begin{figure}
\centering
\psfrag{Sklearn}[c][c][0.7]{\pkg{scikit-learn}}
\psfrag{MRCpy}[c][c][0.7]{\pkg{MRCpy}}
\psfrag{clf = SVM()}[c][c][0.7]{\code{clf1 = SVM()}}
 \psfrag{clf1.fit(...)}[c][c][0.7]{ \ \code{clf1.fit(...)}}
 \psfrag{clf = MRC()}[c][c][0.7]{\code{clf2 = MRC()}}
 \psfrag{clf2.fit(...)}[c][c][0.7]{\code{ \ clf2.fit(...)}}
 \psfrag{clf3.fit(...)}[c][c][0.7]{\color{red} \code{ \ \ \ clf3.fit(...)}}
 \psfrag{...upper()}[c][c][0.7]{\color{red}\code{clf2.get\textunderscore upper\textunderscore bound()}}
  \psfrag{...lower()}[c][c][0.7]{\color{red}\code{clf2.get\textunderscore lower\textunderscore bound()}}
\includegraphics[width=0.8\textwidth]{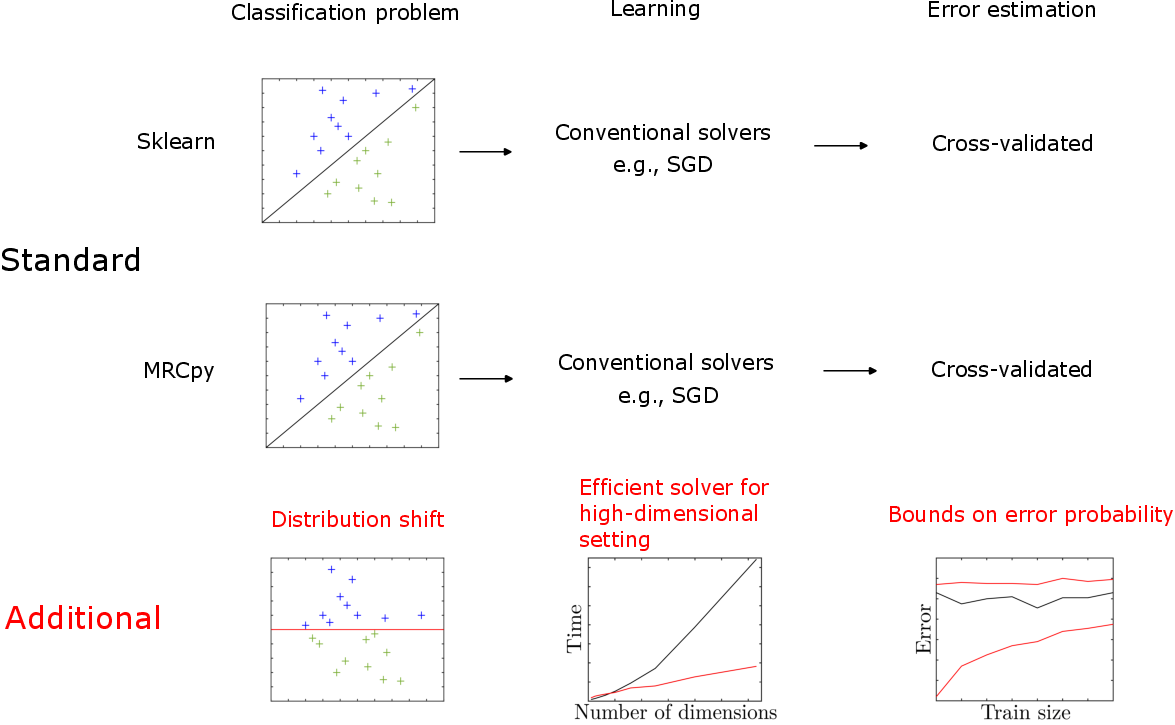}
\captionsetup{font=small}
\caption{The figure shows the functionalities of \pkg{MRCpy} for standard classification problems, in comparison with \pkg{scikit-learn}, together with the additional functionalities provided by \pkg{MRCpy}.}
\label{fig:intro_library_fig}
\end{figure}

\shortcites{PedFab:11, CurEd:23}
Libraries for supervised classification have enabled the wide-spread usage of machine learning methods. Existing libraries, such as \pkg{scikit-learn} in \proglang{Python}, \kar{\pkg{caret} and \pkg{kernlab} in \proglang{R} , and \pkg{mlpack} in \proglang{C++} \citep{PedFab:11, Kuhn:08, kernlab, CurEd:23}}, implement techniques based on the classical \acf{ERM} approach. Such techniques, including \ac{NN}, \ac{SVM}, and \acf{LR}, minimize the risk over the empirical distribution and utilize surrogates to 0-1 loss \citep[see e.g.,][]{MehRos:18}.
Recently, multiple techniques have been proposed following the alternative \acf{RRM} approach (also known as distributionally robust learning), which minimizes the worst-case risk over a set of distributions \citep{AsiXinBeh:15, FatAnq:16}.
\kar{However, very few of the existing libraries implement such approach. We are only aware of the library \pkg{rsome} in \proglang{Python} \cite{rsome} that facilitates the modelling of distributionally robust optimization but does not directly address supervised classification problems}. We present the \proglang{Python} library \pkg{MRCpy} that implements \acfp{MRC} based on the \ac{RRM} approach \citep{MazZanPer:20, MazShePer:22}. \acp{MRC} can utilize \mbox{0-1} loss and provide performance guarantees at learning. Moreover, \pkg{MRCpy} also implements \acp{MRC} that can enable efficient learning in high dimensions \citep{BonMazPer23} and adapt to distribution shifts \citep{AlvMazLoz22, SegMazLiu:23} (see Figure~\ref{fig:intro_library_fig}).

\acp{MRC} minimize the worst-case risk over an uncertainty set of distributions given by linear constraints on the expectations of a feature map. \acp{MRC} can consider various loss functions for the risk, such as 0-1 and log loss \citep{MazShePer:22}, and utilize alternative uncertainty sets that impose additional constraints based on the empirical distribution \citep{MazShePer:22}. The minimization problem of \acp{MRC} is equivalent to a convex optimization problem with L1-regularization which can be efficiently solved using learning algorithms based on subgradient method \citep{MazRomGrun:22} and stochastic gradient method \citep{MehRos:18}. 

\acp{MRC} are specially suitable for classification settings characterised by high-dimensionality or influenced by distribution shifts \citep{BonMazPer23, AlvMazLoz22, alvarez2023minimax,SegMazLiu:23}. These settings are common in multiple classification tasks, including the prediction of cancer based on a large gene expression \citep{GuyIsaEtal:02} or the prediction of electricity prices affected by concept drift \citep{webb:2018}. 
\kar{Currently, several libraries address these problems. For instance, \pkg{mRMR} in \proglang{Python} \citep{DinPen:05} implements feature selection techniques for high-dimensional setting, and \pkg{River} in \proglang{Python}~\citep{montiel2021river} implements online learning methods for concept drift adaptation.}

The presented library implements recent supervised classification techniques called \acp{MRC}. It implements multiple variants of \acp{MRC} with an object-oriented approach that are suitable for different supervised classification settings. The library adheres to the standards of popular machine learning libraries, such as \pkg{scikit-learn}, facilitating readability and easy usage together with a seamless integration with other libraries. Figure~1 summarizes the various additional functionalities of \pkg{MRCpy} in comparison with \pkg{scikit-learn}. \pkg{MRCpy} can be used for standard supervised classification similarly as \pkg{scikit-learn} providing user-friendly interface. Moreover, \pkg{MRCpy} provides additional functionalities for scenarios affected by distribution shifts, efficient solver for high-dimensional settings, and bounds for the error probability. The library is available on GitHub at \url{https://github.com/MachineLearningBCAM/MRCpy} along with a detailed documentation at \url{https://machinelearningbcam.github.io/MRCpy/}.

This paper presents multiple use-cases of the library in addition to basic examples. Specifically, we present an example of usage for efficient hyper-parameter tuning that can be achieved using the performance guarantees provided by \pkg{MRCpy}. \kar{In addition, we present examples to demonstrate how \pkg{MRCpy} can be utilized for supervised classification characterised by high-dimensionality or influenced by distribution shift using multiple real data sets}.

The paper is organized as follows: In Section 2, we present the theoretical background of \acp{MRC}. In Section 3, we present the details of the object-oriented implementation of \acp{MRC} along with a basic example of supervised classification using the library. \kar{Finally, Section 4 presents multiple use cases of the library using real data sets}. 





\section{Theoretical background} \label{sec:theory}
In this section, we describe the theoretical background of the techniques implemented in the library. Firstly, we summarize the framework of \acp{MRC} and then describe \acp{MRC} for classification settings characterised by high-dimensionality or influenced by distribution shift.

\subsection{Minimax risk classification}
\label{sec:mrc}

\subsubsection{Notation}
Supervised classification uses instance-label pairs to determine classification rules that assign labels to instances. We denote by $\mathcal{X}$ and $\mathcal{Y}$ the sets of instances and labels, respectively, with $\mathcal{Y}$ represented by the set $\{1, 2, \ldots, |\mathcal{Y}|\}$. We denote by $\text{T}(\mathcal{X}, \mathcal{Y})$ the set of all classification rules (both randomized and deterministic) and we denote by $\up{h}(y|x)$ the probability with which rule $\up{h} \in \text{T}(\mathcal{X}, \mathcal{Y})$ assigns label $y \in \mathcal{Y}$ to instance $x \in \mathcal{X}$ ($\up{h}(y|x) \in \{0, 1\}$ for deterministic classification rules). In addition, we denote by $\Delta(\mathcal{X}\times\mathcal{Y})$ the set of probability distributions on $\mathcal{X} \times \mathcal{Y}$ and by $\ell(\up{h}, \up{p}) = \mathbb{E}_{\up{p}}\ell(\up{h},(x,y))$ the expected loss of the classification rule $\up{h}\in \text{T}(\mathcal{X}, \mathcal{Y})$ with respect to distribution $\up{p} \in \Delta(\mathcal{X}\times\mathcal{Y})$.

\subsubsection{Minimax risk classifiers}

\acp{MRC} obtain a classification rule \(\up{h}_{\sset{U}}\) that minimizes the maximum expected \(\ell\)-loss with respect to distributions in an uncertainty set \(\mathcal{U}\), that is, 
\begin{equation}
\label{eq:minimax}
\up{h}_{\sset{U}} \in \arg \min_{\up{h}} \max_{\up{p} \in \mathcal{U}} \ell(\up{h},\up{p})
\end{equation}
and, we denote by $\up{R}^*$ the minimax risk against $\set{U}$, i.e.,
\begin{equation}\label{eq_2:Risk_l(U)}
    \up{R}^*=\min_{\up{h}}\max_{\up{p}\in\set{U}}\ell(\up{h},\up{p})
\end{equation}
where $\ell(\up{h},\up{p})$ denotes the expected loss of classification rule $\up{h}$ w.r.t. distribution $\up{p}$.
The uncertainty set considered by the \acp{MRC} are given by constraints on the expectations of a function $\Phi: \set{X} \times \set{Y} \rightarrow \mathbb{R}^{m}$, referred to as a feature mapping, as 
\begin{equation}
\label{eq:mrc_uncertainty_set}
\set{U} = \{\up{p} \in \Delta(\set{X} \times \set{Y}): |\mathbb{E}_{\up{p}}\{\Phi\} - \B{\tau}| \preceq \B{\lambda}\},
\end{equation}
where |·| denotes the vector formed by the absolute value of each component in the argument, $\B{\tau}$ denotes the mean vector of expectation estimates corresponding with $\Phi$, and $\B{\lambda} \succeq 0$ is a confidence vector that accounts for inaccuracies in the estimate. \acp{MRC} using such type of uncertainty set can provide bounds on the expected loss as detailed in Section~\ref{sec:mrc_learning}.

The uncertainty set can also include an additional constraint that fixes the instances' marginal distribution $\up{p}_x$ with the empirical marginal distribution $\up{p}_{x}^{n}$ as 
\begin{equation}
\label{eq:cmrc_uncertainty_set}
\set{V} = \{\up{p} \in \Delta(\set{X} \times \set{Y}): |\mathbb{E}_{\up{p}}\{\Phi\} - \B{\tau}| \preceq \B{\lambda} \ \up{and} \ \up{p}_x = \up{p}_{x}^{n}\}.
\end{equation}
\acp{MRC} using this type of uncertainty set can correspond to popular techniques, such as L1-regularized \ac{LR}.

The feature mappings $\Phi: \set{X}\times \set{Y} \rightarrow \mathbb{R}^m$ used by \acp{MRC} represent \mbox{instance-label} pairs as real vectors similarly to other existing techniques \citep[see e.g.,][]{MehRos:18}. The most common way to define such feature mapping is using multiple features over instances together with one-hot encodings of labels as follows
\begin{equation}
\label{eq:feature}
 \Phi(x, y)  = \, \V{e}_{y} \otimes {\Psi}(x)=  \big{[}\mathds{1}\{y = 1\} {\Psi}(x)^{\top}, \mathds{1}\{y = 2\} {\Psi}(x)^{\top}, \ldots, \mathds{1}\{y = |\mathcal{Y}|\} {\Psi}(x)^{\top}\big{]}^{\top}
\end{equation}
where $\otimes$ denotes the Kronecker product, $\B{\up{e}}_i$ denotes the i-th vector in a standard basis, $\mathds{1} \{\cdot\}$ denotes the indicator function, and the map $\Psi : \mathcal{X} \rightarrow \mathbb{R}^d$ represents instances as real vectors of size $d$ . This map can be just the identity $\Psi(x)=x$, multiple polynomials on $x$, or the last layer of a \ac{NN} \citep{BenCouAar:13}. 
 
The mean vector $\B{\tau} = [\tau^{(1)}, \tau^{(2)}, \ldots, \tau^{(m)}]^{\top}$ in \eqref{eq:mrc_uncertainty_set} and \eqref{eq:cmrc_uncertainty_set} is an estimate of the feature mapping expectation $\mathbb{E}_{\up{p}^{*}}\{\Phi\}$ with respect to the true underlying distribution $\up{p}^{*}$. The confidence vector $\B{\lambda} = [\lambda^{(1)}, \lambda^{(2)}, \ldots, \lambda^{(m)}]^{\top}$ is an estimate of the mean vector accuracy, and controls the size of the uncertainty set considered. These vectors can be obtained from instance-label pairs $(x_1,y_1), (x_2, y_2), \dots, (x_n, y_n)$ as
\begin{equation}
\label{eq:estimates}
\tau^{(j)} = \frac{1}{n} \sum_{i=1}^{n} \Phi^{(j)}(x_i, y_i), \ \lambda^{(j)}=\sqrt{\frac{\sum_{i=1}^{n}(\Phi^{(j)}(x_i, y_i) - \tau^{(j)})^{2}}{n^{2}}} \ \text{for} \ j=1,2,\ldots,m.
\end{equation}
where $\Phi^{(j)}(x,y)$ is a component of the feature mapping in \eqref{eq:feature}.

The expected loss given by $\ell(\up{h},\up{p}) = \mathbb{E}_{\up{p}}\ell(\up{h},(x,y))$ for rule $\up{h}$ quantifies the classification risk with respect to distribution $\up{p} \in \Delta(\set{X} \times \set{Y})$. \acp{MRC} can be implemented using general loss functions \citep{MazShePer:22} for instance \mbox{0-1} loss that is given by \mbox{$\ell_{01}(\up{h},(x,y)) = 1-\up{h}(y|x)$}, and log loss that is given by $\ell_{\text{log}}(\up{h},(x,y)) = -\log \up{h}(y|x)$. The \mbox{0-1} loss is particularly suitable for discriminative approaches since it quantifies the classification error, while the log loss can be more suitable for conditional probability estimation since it scores probability assessments.

\subsubsection{MRCs learning}
\label{sec:mrc_learning}
\noindent
The minimax risk optimization problem \eqref{eq:minimax} composes the learning stage of \acp{MRC} and is equivalent to a convex optimization problem with \mbox{L1-regularization} \citep{MazShePer:22}. For instance, the \ac{MRC} rule $\up{h}_{\sset{U}}$ solution of \eqref{eq:minimax} corresponding to 0-1 loss is given by a coefficient vector $\B{\mu}^{*} \in \mathbb{R}^{m}$ obtained by solving the following optimization problem \citep{MazRomGrun:22}
\begin{equation}
\begin{gathered}
\label{eq:mrc_opt}
\up{R}^{*} = \underset{\B{\mu}}{\min} \; 1 - \B{\tau}^{\top} \B{\mu} + \B{\lambda}^{\top} | \B{\mu} | + \max_{x \in \set{X}, \set{C} \subseteq \set{Y}} \left(\frac{\sum_{y \in \set{C}} \Phi(x,y)^{\top}\B{\mu} - 1}{|\set{C}|}\right).
\end{gathered}
\end{equation}

The \ac{MRC} rule $\up{h}_{\sset{U}}$ given by $\B{\mu}^{*}$ assigns label $y \in \set{Y}$ to instance $x \in \set{X}$ based on probability $\up{h}_{\sset{U}}(y|x)$ as detailed in \cite{MazRomGrun:22}, and the corresponding deterministic rule $\up{h}^{\up{d}}_{\sset{U}}$ assigns labels with maximum probability $\up{h}_{\sset{U}}(y|x)$, that is, \mbox{${\arg} \max_{y \in \set{Y}} \up{h}_{\sset{U}}(y|x)=\arg \max_{y \in \set{Y}}\Phi(x,y)^\top\B{\mu}^*$}.  Moreover, the classification risk $\mathcal{R}(\up{h}_{\sset{U}})$ of the rule $\up{h}_{\sset{U}}$ is upper bounded by the following inequality
\begin{equation}
\label{eq:minimax_risk_bound}
\mathcal{R}(\up{h}_{\sset{U}}) \leq \up{R}^{*} + (|\mathbb{E}_{\up{p}^{*}}\{\Phi(x,y)\}-\B{\tau}|-\B{\lambda})^{\top}|\B{\mu}^{*}|.
\end{equation}
In particular, if $\B{\lambda}$ is a confidence vector with coverage probability $1-\delta$, that is, $\mathbb{P}\{|\mathbb{E}_{\up{p}^{*}}\{\Phi(x,y)\}-\B{\tau}| \preceq \B{\lambda}\} \geq 1-\delta$, then $\mathcal{R}(\up{h}_{\sset{U}})  \leq \up{R}^{*}$ with probability at least $1-\delta$. In addition, the lower bound on the classification risk $\mathcal{R}(\up{h}_{\sset{U}})$ is obtained by solving a related convex optimization problem defined in \cite{MazRomGrun:22}.

\acp{MRC} using uncertainty sets $\set{U}$ and $\set{V}$ with different loss functions lead to different convex optimization problems \citep{MazShePer:22}. All these problems can be solved using generic algorithms for convex optimization, such as those implemented in \pkg{CVXPY} \citep{DiaBoy:16} which can be highly precise but computationally expensive. In the following, we describe different methods that can efficiently solve the optimization problem of particular \acp{MRC} while being less accurate than \pkg{CVXPY}.

\subsubsection{Subgradient method}
Subgradient method is an attractive option to solve non-differentiable convex optimization. \cite{MazRomGrun:22} presents an algorithm based on the subgradient method to enable fast optimization of the \acp{MRC} corresponding to uncertainty set $\set{U}$ and using \mbox{0-1} loss \eqref{eq:mrc_opt}. In particular, the algorithm utilizes accelerated subgradient methods based on Nesterov's extrapolation \citep{TaoPanWuTao:2020} that have been developed to reduce the number of iterations.

\subsubsection{Constraint generation method}
Constraint generation method \citep{DimTsi97} is suitable for solving \mbox{large-scale} \acp{LP}, such as the \ac{LP} formulation of \acp{MRC} corresponding to uncertainty set $\set{U}$ and 0-1 loss given in \cite{BonMazPer23}. The paper presents an algorithm based on constraint generation that enables efficient learning of \acp{MRC} in settings with large number of features in the data. Particularly, the algorithm leverages the sparsity induced by the \mbox{L1-penalization} in the optimization problem \eqref{eq:mrc_opt} and the fact that usually only a small subset of features are informative in high-dimensional settings \citep{GhoDeb:22}.

\subsubsection{Stochastic gradient descent}
Stochastic gradient descent \citep{MehRos:18} is a popular method for solving convex optimization problems. The library implements the \ac{SGD} method and its Adam variant \citep{KingBa15} for \acp{MRC} corresponding to uncertainty set $\set{V}$.
\newline
\newline
The following sections present multiple \acp{MRC} that are specially suitable for classification settings characterised by high-dimensionality or influenced by distribution shifts.

\subsection{MRCs for high-dimensional data}\label{sec:high_dimensions}
High-dimensional data is common in multiple areas such as health care and genomics. A typical example is to classify patients as healthy or having cancer based on gene expression data with tens of thousands of features \citep{GuyIsaEtal:02}. In addition to high-dimensional raw data, the feature mapping $\Phi$ often performs a high-dimensional representation of the input data vector in order to improve the classification performance. Learning in such high-dimensional setting often leads to highly complex optimization processes because the number of variables involved in the optimization increases with the number of features \citep{ShiYin:10, YuaGuo:12}. In addition to high complexity, the conventional performance assessment based on cross-validation \citep{rodriguez2013general} can be unreliable with limited number of instance-label pairs in the above-mentioned applications, e.g., tens of thousands of features but less than 100 patients \citep{GuyEli:03, BroGavEtal:12}.

\cite{BonMazPer23} presents an efficient learning algorithm for the optimization problem \eqref{eq:mrc_opt} of \acp{MRC} in high-dimensional settings. In particular, the algorithm based on constraint generation method enables efficient learning of the \ac{LP} formulation of \eqref{eq:mrc_opt} presented in \cite{BonMazPer23}. The efficient learning is feasible due to the fact that the \ac{MRC}'s solution is a sparse vector in high dimensions. This sparsity is induced by L1-regularization in \eqref{eq:mrc_opt}, and the fact that usually only a small subset of features are informative in high-dimensional settings.

The algorithm obtains the optimal solution to \eqref{eq:mrc_opt} by iteratively solving a sequence of \ac{MRC} subproblems. These subproblems are given by a subset of features selected by the constraint generation approach. The number of features selected in each iteration is restricted by hyper-parameter $n_\up{max}$ to control the complexity in each iteration. Decreasing $n_{\up{max}}$ decreases the complexity per iteration while increasing the number of iterations to obtain the optimal solution. \cite{BonMazPer23} shows that the optimal solution $\up{R}^{k}$ obtained by the algorithm satisfies
\begin{equation}\label{ineq:convergence}
\up{R}^{*} \leq \up{R}^{k} \leq \up{R}^{*} + \epsilon\|\B{\mu}^{*}\|_1,
\end{equation}
where $\up{R}^{*}$ and $\B{\mu}^{*}$ are solution obtained by solving \eqref{eq:mrc_opt} using all the features and $\epsilon$ is a \mbox{hyper-parameter} of the algorithm that controls the accuracy of the solution obtained.

\subsection{MRCs for supervised classification under concept drift}\label{sec:AMRC}

The statistical characteristics describing the underlying distribution of instance-label pairs often change with time in practical scenarios of supervised classification \citep{gama:2014}. Such concept drift is common in multiple applications including electricity price prediction \citep{webb:2018} and spam mail filtering \citep{delany:2005}. For instance, in the problem of predicting electricity price increases/decreases, the statistical characteristics related to electricity demand, generation, and price often change over time due to varying habits and weather. 

Supervised classification techniques adapt to concept drift by updating classification rules as new instance-label pairs arrive. Conventional learning techniques account for a scalar rate of change by means of a carefully chosen parameter such as a learning rate \citep{orabona:2008}, forgetting factor \citep{pavlidis:2011}, or window size \citep{nguyen2018variational}. However, in common scenarios, the concept drift cannot be adequately addressed accounting only for a scalar rate of change. Such inadequacy is due to the fact that time changes are commonly multidimensional, i.e., different statistical characteristics of instance-label pairs often change in a different manner. For instance, in the problem of predicting electricity price increases/decreases, the statistical characteristics related to demand often change differently from those related to generation. 

\cite{AlvMazLoz22} presents \acp{AMRC} that account for multidimensional time changes and can provide tight performance guarantees. The learning methodology of \ac{AMRC} provides multidimensional adaptation by estimating multiple statistical characteristics of the time-varying underlying distribution; and can provide computable tight performance guarantees under concept drift in terms of instantaneous bounds for error probabilities and accumulated mistakes.

The methodology of \acp{AMRC} presented in~\cite{AlvMazLoz22} sequentially obtains a 0-1 \ac{MRC} rule associated with the uncertainty set $\set{U}$ every time a new instance-label pair is received. At each time $t$, \acp{AMRC} obtain an updated rule $\up{h}_t$ by recursively updating mean vector $\B{\tau}_t$ and the confidence vector $\B{\lambda}_t$, and efficiently solving the optimization problem~\eqref{eq:mrc_opt}; and provide performance guarantees by using the minimax risk $\up{R}^{*}_t$.

\subsection{MRCs for supervised classification under general covariate shift}\label{Sec:DWGCS}
Covariate shift in supervised classification refers to the scenarios in which the marginal distributions of instances (covariates $x$) at training and testing are different, $\ptr(x)$ and $\pte(x)$ respectively, while the conditional distribution over the labels remains the same \citep{SugKaw:12,QuiSugSch:08}. 
Such scenarios are common for classification in, for instance, medical applications, such as electronic health record data analysis \citep{SinMhaChu:22}, where the patients used as training population belong to different hospitals. 
This may be explained, among other reasons, by the difficulty in obtaining data from patients in the same hospital or the unavailability of such data. 

Common covariate shift adaptation techniques are based on reweighted approach \citep{SugKaw:12,QuiSugSch:08,CorMohRil:08,Zad:04} that weight loss functions at training using the ratio $\pte(x)/\ptr(x)$. 
Such reweighted techniques assign higher weights to the training instances that are more likely at testing.
Reweighted techniques are designed for situations where the support of the training distribution contains the support of the testing distribution. 
However, even when the support condition is satisfied, reweighted approaches may achieve poor performance if the ratios described above take large values, leading to poor expectation estimates.
On the other hand, robust approaches \citep{CheMonLiu:16,LiuZie:14} weight feature mappings at testing using the ratio $\ptr(x)/\pte(x)$.
Such robust techniques assign low confidence to the testing instances that are unlikely at training. 
Robust techniques are designed for situations where the support of the testing distribution contains the support of the training distribution. 
However, even when the support condition is satisfied, robust approaches may achieve poor performance if the ratios described above take large values, leading to overconfident classification rules.

\cite{SegMazLiu:23} presents \ac{DW-GCS} approach that adapts to covariate shift without any prior knowledge over the supports over the training and testing distributions. 
The learning methodology proposed in \cite{SegMazLiu:23} tackles the general covariate shift by weighting both training and testing instances by $\beta(x)$ and $\alpha(x)$ respectively, such that $\ptr(x)\beta(x)=\pte(x)\alpha(x)$; and provides generalization bounds that show a significant increase in the effective sample size in comparison with existing reweighted approaches.

The methodology of \ac{DW-GCS} presented in \cite{SegMazLiu:23} obtains \ac{MRC} rules for both 0-1 and log loss associated with the uncertainty set $\set{V}$ defined in terms of constrains on the expectation of the weighted feature mapping $\Phi_{\alpha}(x,y)=\alpha(x)\Phi(x,y)$.
The expectation can be estimated using averages of training instances weighted by $\beta(x)$ since $\mathbb{E}_{\pte}\alpha(x)\Phi(x,y)=\mathbb{E}_{\ptr}\beta(x)\Phi(x,y)$. 
Weights $\beta(x)$ and $\alpha(x)$ are obtained generalizing the conventional \ac{KMM} technique \citep{HuaSmoGre:06}.



\section{Implementation and usage of MRCpy} \label{sec:design}

In this section, we detail the library's implementation of the \acp{MRC} described in Section~\ref{sec:theory} along with a basic usage example. \kar{The library is hosted on GitHub at \url{https://github.com/MachineLearningBCAM/MRCpy} and can be installed via the \pkg{Python} Package Index (PyPI) using}
\begin{Code}
pip install MRCpy
\end{Code}
\kar{or with the readers' favorite \pkg{Python} package installation program.

\subsection{Extendable architecture of MRCpy}
\label{subsec:arch_mrcpy}
\pkg{MRCpy} provides implementation for multiple variants of \acp{MRC} following the standards of popular machine learning libraries, such as \pkg{scikit-learn}, to facilitate readability and the seamless integration with other libraries. In particular, \pkg{MRCpy} uses an object-oriented approach that can be easily extended to implement multiple variants of \acp{MRC} corresponding to different uncertainty sets and loss functions as described in the following.

Each \ac{MRC} class inherits from a skeleton class \class{BaseMRC} and has the following attributes that correspond to the descriptions in Section~\ref{sec:mrc}.
\begin{itemize}
  \item \code{phi} (type: string, values: \{\code{"linear"}, \code{"fourier"}, \code{"relu"}, \code{"threshold"}\}): indicates the type of feature mapping $\Phi$. Currently, the library implements features $\Psi(x)$ based on Fourier \citep{RahRec:08}, ReLU \citep{SunGilTew:19}, and threshold \citep{LebLaf:01, MazZanPer:20} in addition to the usual linear feature map, that is, $\Psi(x)=x$.
  
  \item \code{loss} (type: string, values: \{\code{"0-1"}, \code{"log"}\}): indicates the type of loss function $\ell$. 
  
  \item \code{deterministic} (type: boolean): indicates whether the classification rule is deterministic or not.

  \end{itemize}
In addition, each \ac{MRC} class inheriting from class \class{BaseMRC} implements the functions 
\begin{itemize}
    \item \code{minimax\textunderscore risk()}: solves the \ac{MRC} optimization corresponding to an uncertainty set and loss function, and obtains the model parameters.
    \item \code{predict\textunderscore proba()}: computes the prediction probabilities for given testing samples using the model parameters.
\end{itemize}
The class \class{BaseMRC} implements the common functionalities through the functions
\begin{itemize}
    \item \code{fit()}: computes the feature mappping $\B{\Phi}$, $\B{\tau}$, and $\B{\lambda}$ as defined in Section~\ref{sec:mrc}, and obtains the model parameters using the subclass function \code{minimax\textunderscore risk()}.
    \item \code{predict()}: assigns the deterministic/non-deterministic label using on the probabilities given by the subclass function \code{predict\textunderscore proba()}.
\end{itemize}

\pkg{MRCpy} also enables to easily implement feature mappings and combine them with the \acp{MRC} in the library. The module \code{phi} of the library provides the implementation of different kinds of feature mappings with an object-oriented approach. A feature mapping in the module inherits from a class \class{BasePhi}, and implements the functions \code{fit()} and \code{transform()}. The function \code{fit()} learns the required parameters used to compute the features and the function \code{transform()} returns the features computed using the parameters.

In the following, we present the details of multiple \acp{MRC} implemented in the library along with basic example usage for supervised classification. In addition, Section~\ref{sec:numerical_ex} presents additional examples for multiple use-cases of the library. Detailed documentation of the library along with multiple examples can also be found at \url{https://machinelearningbcam.github.io/MRCpy/}.}

\subsection{MRCpy for standard supervised classification}\label{sec:basic_example}
\kar{
The classes \class{MRC} and \class{CMRC} in \pkg{MRCpy} implements techniques for standard supervised classification as described in Section~\ref{sec:mrc}. In particular, the classes implement the \acp{MRC} corresponding to uncertainty sets $\set{U}$ and $\set{V}$ as defined in \eqref{eq:mrc_uncertainty_set} and \eqref{eq:cmrc_uncertainty_set}, respectively where each class provides implementation for both \mbox{0-1} and \mbox{log} loss. 

Each of these classes can implement multiple variations using the attributes defined in the Section~\ref{subsec:arch_mrcpy}. The default values of the attributes in these classes are \code{phi = "linear"}, \code{loss = "0-1"}, and \mbox{\code{deterministic = True}}. In addition, the class attribute \code{solver} indicates the optimization method used for learning the \acp{MRC} as discussed in Section~\ref{sec:mrc_learning}. Table~\ref{tb:solvers} summarizes the \code{solver} options available for \class{MRC} and \class{CMRC} classifier based on the loss function. The default values for \class{MRC} and \class{CMRC} classifiers are \code{solver = "subgrad"} (subgradient method) and \code{solver = "adam"} (Adam variant of \ac{SGD}), respectively. 

In the following, we present an example that illustrates the usage of the library for standard supervised classification using real data set. The example is presented for \class{MRC} classifier using \mbox{0-1} loss that corresponds to the optimization problem \eqref{eq:mrc_opt} in Section~\ref{sec:mrc_learning}.
}
\begin{table}
\small
\def\arraystretch{1.5}
\renewcommand{\arraystretch}{2}
\centering
\caption{Available solvers for \acp{MRC} in \pkg{MRCpy}}
\begin{tabular}{ccccc}
\hline
 \multirow{2}{*}{\code{solver}} 	&  \multicolumn{2}{c}{\class{MRC}} & \multicolumn{2}{c}{\class{CMRC}} \\
& \mbox{0-1} & log & \mbox{0-1} & log \\
 \hline
 \hline
 \code{cvx} 		& {\color{teal}\checkmark} & {\color{teal}\checkmark} & {\color{teal}\checkmark} & {\color{teal}\checkmark} \\
 \hline
 \code{subgrad} 	& {\color{teal}\checkmark} & {\color{teal}\checkmark} & {\color{red}\xmark} & {\color{red}\xmark} \\
 \hline
 \code{sgd} 		& {\color{red}\xmark}  & {\color{red}\xmark}  & {\color{teal}\checkmark} & {\color{teal}\checkmark} \\
 \hline
 \code{adam} 		& {\color{red}\xmark} & {\color{red}\xmark} & {\color{teal}\checkmark} & {\color{teal}\checkmark} \\
 \hline
 \code{cg} & {\color{teal}\checkmark} & {\color{red}\xmark} & {\color{red}\xmark}  & {\color{red}\xmark} \\
 \hline
\end{tabular}
\label{tb:solvers}
\end{table}

\subsubsection{Example: Supervised classification}
To start off, we load the ``indian\textunderscore liver'' real data set available in the \code{MRCpy.datasets} module of the library. To load the data, we first import the function \code{load_indian_liver()} from \code{MRCpy.datasets} module and then call that function to load the data as \pkg{NumPy} \citep{Oli:06} matrices :
\begin{Code}
>>> from MRCpy.datasets import load_indian_liver
>>> X, Y = load_indian_liver()
\end{Code}
After the data is loaded, we split the data into training and testing set using the \pkg{scikit-learn} library as
\newline
\newline
\code{>>> from sklearn.model_selection import train_test_split}
\newline
\code{>>> X_train, X_test, Y_train, Y_test = train_test_split(X, Y, }
\newline
\code{... test_size = 0.2, random_state = 1)}
\newline
\newline
Now, import the \class{MRC} class from the library into a working Python environment and define a \class{MRC} classifier instance \code{clf} using the default attributes as
\newline
\newline
\code{>>> from MRCpy import MRC}
\newline
\code{>>> clf = MRC()}
\newline
\newline
Then, the \code{clf} classifier can be trained using its \code{fit()} function that solves the \ac{MRC} optimization problem \eqref{eq:mrc_opt} using the estimates $\B{\tau}$ and $\B{\lambda}$ defined in \eqref{eq:estimates}: 
\newline
\newline
\code{>>> clf.fit(X_train, Y_train)}
\newline
\newline
After fitting the classifier \code{clf}, prediction can be done using its \code{predict()} function and the error can be obtained using its \code{error()} functions as
\newline
\newline
\code{>>> print("Predicted labels for 10 instances: ", clf.predict(X_test)[:10])}
\newline
\code{>>> print("Classification error: ", clf.error(X_test, Y_test))}
\begin{CodeOutput}
Predicted labels for 10 instances: array([0, 0, 0, 0, 0, 0, 0, 0, 0, 0])
Classification error: 0.2905982905982906
\end{CodeOutput}
\subsubsection{Example: Bounds on the classification error}
The function \code{get_upper_bound()} of the \class{MRC} classifier returns the upper bound on the classification error. In case of non-deterministic classification, the upper bound $\up{R}^{*}$ is obtained as a result of fitting the classifier (solving \eqref{eq:mrc_opt}). In case of deterministic classification, the upper bound $\up{R}^{*}$ is obtained by solving a related convex optimization problem given in \cite{MazRomGrun:22}. 

The function \code{get_lower_bound()} of the \class{MRC} classifier returns the lower bound on the classification error. The lower bound is obtained by solving an additional optimization problem for both deterministic and non-deterministic classification \citep{MazRomGrun:22}. 

For the previous example, the bounds on the classification error of the \code{clf} classifier are:  
\newline
\newline
\code{>>> print("Upper bound on the error: ", clf.get_upper_bound())}
\newline
\code{>>> print("Lower bound on the error: ", clf.get_lower_bound())}
\begin{CodeOutput}
Upper bound on the error: 0.291683832000348
Lower bound on the error: 0.27913161864342884
\end{CodeOutput}
\subsection{MRCpy for non standard supervised classification }
In this subsection, we detail the implementation of the \acp{MRC} for specific classification settings discussed in Section 2.

\subsubsection{MRCpy for high-dimensional data}
The class \class{MRC} implements the learning algorithm presented in \cite{BonMazPer23} for \mbox{0-1} loss to enable efficiency in high dimensions. The algorithm can be used by setting the class attribute \code{solver = "cg"} which refers to the constraint generation approach as described in Section~\ref{sec:high_dimensions}. The training time and the accuracy of the algorithm can be controlled using the class attributes \code{n_max} and \code{eps} respectively. The default values are \code{n_max = 100} and \code{eps = 0.0001}. 

\subsubsection{MRCpy for concept drift adaptation}\label{sec:AMRC_implementation}
\kar{The class \class{AMRC} in \pkg{MRCpy} library implements techniques presented in~\cite{AlvMazLoz22} for supervised classification under concept drift. In particular, the class implements the \ac{MRC} rule associated with uncertainty set $\set{U}$ and \mbox{0-1} loss at each time step as described in Section~\ref{sec:AMRC}.

The \class{AMRC} class can implement multiple variations using the attributes defined in the Section~\ref{subsec:arch_mrcpy}. The default values for these attributes are same as for class \class{MRC}. However, the class attribute \code{phi} can only be set to \code{linear} or \code{fourier} and the attribute \code{loss} can only be set to \code{0-1} as \acp{AMRC} are not defined for log loss. The \class{AMRC} class also implements the following functions
\begin{itemize}
    \item \code{get\_upper\_bound()}: returns the upper bound on the expected loss for the fitted model. Specifically, for the instance at time $t$, this function returns the minimax risk $\up{R}_t^*$ that bounds the error probability of the classification rule.
    \item \code{get\_upper\_bound\_accumulated()}: returns the upper bound on the accumulated mistakes of the fitted model. Specifically, at time $t$, this function returns \mbox{$\sum_{i = 1}^t \up{R}_i^*/t +\sqrt{2 \log(1/\delta)/t}$} that bounds the accumulated mistakes per time step with probability at least $1-\delta$. The default value for $\delta$ is $0.05$.
\end{itemize}
Note that the class attribute \code{delta} corresponds to $\delta$ that defines the confidence of the upper bound on the accumulated mistakes.
}

\subsubsection{MRCpy for general covariate shift adaptation}
The class \class{DWGCS} in the library implements the \ac{DW-GCS} presented in~\cite{SegMazLiu:23} for supervised classification under general covariate shift. \kar{In particular, 
 the class \class{DWGCS} obtains \ac{MRC} rules associated with uncertainty set $\set{V}$ defined in terms of the weighted feature mapping $\Phi_{\alpha}$ using both 0-1 and log loss as described in Section~\ref{Sec:DWGCS}.
 
The \class{DWGCS} class can implement multiple variations using the attributes defined in Section~\ref{subsec:arch_mrcpy}. The default values for these attributes are same as for class \class{CMRC}. However, the class attribute \code{phi} can only be set to \code{linear}, \code{relu}, or \code{fourier}. In addition, the class attributes \code{weight\_alpha} and \code{weight\_beta} allow the user to manually enter the weights $\alpha(x)$ and $\beta(x)$ (see details in Section~\ref{Sec:DWGCS}).
If none of the weights are passed, they are computed by solving the optimization problem \ac{DW-KMM} defined in Section~4 of \cite{SegMazLiu:23}.
The class attribute \code{D} also controls the values of the weight functions $\alpha(x)$ and $\beta(x)$, and corresponds to the hyperparamter $D$ described in Section~4 of \cite{SegMazLiu:23}.
\code{D = 1} implements conventional \ac{KMM}, and \code{D = np.inf} sets the weights $\beta(x)=1$ and obtain weights $\alpha(x)$ by solving the conventional \ac{KMM} method \citep{HuaSmoGre:06}.
The default value of attribute \mbox{\code{D = 4}}.}



\section{Examples and results} \label{sec:numerical_ex}
\kar{In this section, we present examples of usage along with numerical results for multiple use-cases of the library using publicly available real datasets summarized in Table~\ref{tb:datasets}}. Firstly, we present example of usage for efficient hyper-parameter tuning that can be achieved using the performance guarantees provided by \pkg{MRCpy}. Then, we present examples of usage for specific classification settings discussed in Section~\ref{sec:theory}. \kar{Note that the parameter \code{random\textunderscore state} in all the code examples corresponds to the random seed. We fix this parameter in the following examples to ensure that the results are reproducible.}

\begin{table}
 \captionsetup{labelfont={it}, labelsep=period, font=small}
                         \caption{Real binary classification data sets.}
    \vskip -0.15in
     \label{tb:datasets}
\setstretch{1.2}
\begin{center}
\scalebox{0.85}{\begin{tabular}{|c|c|c|c|c|}
\hline
Data set & \multicolumn{1}{c|}{Samples} & \multicolumn{1}{c|}{Dimensions} & Classification type \\ \hline
Mammographic & 961 & 14 & standard \\
Haberman & 306 & 3 & standard \\
Indian\textunderscore liver & 583 & 10 & standard \\
Diabetes & 768 & 8 & standard \\
Colon & 62 & 2000 & high-dimensional \\
Leukemia & 72 & 7129 & high-dimensional \\
Ovarian & 253 & 15154 & high-dimensional \\
Prostate\textunderscore GE & 102 & 12600 & high-dimensional \\
Usenet2 & 1,500 & 100 & concept drift \\
Agrawal-abrupt (synthetic) & 1,000,000 & 9 & concept drift \\
Airlines  & 539,383 & 7 & concept drift \\
SMTP & 95,156 & 3 & concept drift \\
HTTP & 567,498 & 3 & concept drift \\
Credit & 690 & 15 & concept drift \\
20 Newsgroups & 18846 & 61188 & covariate shift \\

 \hline
\end{tabular}}
\end{center}
   \vskip -0.25in
\end{table}

\subsection{Hyper-parameter tuning: upper bound vs cross-validation error}
In the following, we illustrate the usage of \pkg{MRCpy} for efficient hyper-parameter tuning. In particular, we use the upper bound given by \pkg{MRCpy} for hyper-parameter tuning and compare it with the usual 10-fold cross-validation error approach implemented by \code{RandomizedSearchCV()} function in \pkg{scikit-learn}. \kar{The results using multiple real data sets show that the hyper-parameter tuning based on \ac{MRC}'s upper bound can be 10 times faster than the cross-validation approach while obtaining classifiers with similar performance.}

The following code selects the scaling parameter of Fourier features for \acp{MRC} using the usual cross-validation approach and the approach based on upper bound.

\subsubsection{Loading the data set}
To start off, we load the ``haberman'' data set from the \code{MRCpy.datasets} module of the library:
\newline
\newline
\code{>>> from MRCpy.datasets import load_haberman}
\newline
\code{>>> X, Y = load_haberman()}
\newline
\newline
Then, we split the data into training and testing set. The training set is used for parameter selection and the testing set is used to estimate the error of the model trained using the selected parameter. The \code{train_test_split()} function of \pkg{scikit-learn} library is used to obtain split:
\newline
\newline
\code{>>> from sklearn.model_selection import RandomizedSearchCV, train_test_split}
\newline
\code{>>> X_train, X_test, Y_train, Y_test = \textbackslash}  
\newline
\code{... train_test_split(X, Y, test_size = 0.2, random_state = 1)}
\newline
\newline
Now, we create 10 instances of the Fourier feature mapping class \class{RandomFourierPhi} of \pkg{MRCpy} which are defined by different values for the scaling parameter \code{sigma} as follows:
\begin{Code}
>>> from MRCpy.phi import RandomFourierPhi
>>> n_iter = 10
>>> d = X.shape[1]
>>> phi_arr = []
>>> for i in range(1, n_iter + 1):
...    phi_arr.append(RandomFourierPhi(n_classes = 2,
...    sigma = (d / (i + 1)), random_state = 42))
\end{Code}
In the following, we use the different instances of Fourier features in \code{phi_arr} to select the \code{phi} attribute of the \class{MRC} class that defines the feature mapping.

\subsubsection{Tuning using cross-validation}
In this part, we use the cross-validation approach to select \code{phi}. The following code snippet uses the \class{RandomizedSearchCV} class of \pkg{scikit-learn} library to implement this approach. The object requires the attributes: the classifier and the set of possible values for \code{phi}. We use the \class{MRC} classifier and the dictionary \code{param} that encloses the possible values for the \mbox{hyper-parameter} \code{s}. Then, its \code{fit()} function uses the training set \code{X_train} and \code{Y_train} to obtain the 10-fold cross-validation error for each value, and the value with minimum cross-validation error is chosen for the estimator:
\begin{Code}
>>> startTime = time.time()
>>> param = {"phi": phi_arr}
>>> mrc = MRC(random_state = 42, deterministic = False, solver = "subgrad")
>>> clf = RandomizedSearchCV(mrc, param, cv = 10, random_state = 42,
... n_iter = n_iter)
>>> clf.fit(X_train, Y_train)
>>> print("Total time taken (in seconds): ", time.time() - startTime)
>>> print("Classification error: ", 1 - clf.score(X_test, Y_test))
\end{Code}
\begin{CodeOutput}
Total time taken (in seconds): 51.678471326828
Classification error: 0.30645161290322576
\end{CodeOutput}

\subsubsection{Tuning using upper bound}
Now, we use the upper bound given by \class{MRC} classifier to select \code{phi}. Firstly, we import the \class{MRC} classifier from \pkg{MRCpy}, and the libraries \pkg{time} and \pkg{numpy}:
\newline
\newline
\code{>>> from MRCpy import MRC}
\newline
\code{>>> import time}
\newline
\code{>>> import numpy as np}
\newline
\newline
We use the \pkg{time} library to compare the computation time of the two methods for hyper-parameter tuning. 

In the following code snippet, we train the \class{MRC} classifier over the whole training set \code{X_train} and \code{Y_train} for each instance in \code{phi_arr}. The upper bound on the classification error corresponding to each value is saved in \code{upps} array.
\begin{Code}
>>> startTime = time.time()
>>> upps = np.zeros(n_iter)
>>> for i in range(n_iter):
...    clf = MRC(phi = phi_arr[i], random_state = 42, deterministic = False,
...    solver = "subgrad")
...    clf.fit(X_train, Y_train)
...    upps[i] = clf.get_upper_bound()
>>> print("Total time taken (in seconds): ", time.time() - startTime)
\end{Code}
\begin{CodeOutput}
Total time taken (in seconds): 5.186525821685791
\end{CodeOutput}
Then, we choose the feature mapping \code{phi} with the minimum upper bound on the classification error:
\begin{Code}
>>> min_upp = np.min(upps)
>>> best_phi = phi_arr[np.argmin(upps)]
\end{Code}
The classification error for the chosen feature mapping \code{best_phi} on the test data \code{X_test} is obtained as:
\begin{Code}
>>> clf = MRC(phi = best_phi, random_state = 42, deterministic = False,
... solver = "subgrad")
>>> clf.fit(X_train, Y_train)
>>> print("Classification error: ", clf.error(X_test, Y_test))
\end{Code}
\begin{CodeOutput}
Classification error: 0.3064516129032258    
\end{CodeOutput}
The code example above shows that the hyper-parameter tuning approach based on \ac{MRC}'s upper bound can be around 10 times faster than the usual 10-fold cross-validation approach with similar performance in terms of accuracy.

Table~\ref{tb:hyper_parameter_tuning} presents results using multiple data sets to compare the performances of the both the above approaches. The results are averaged over different partitions of each data set.

\begin{table}
 \captionsetup{labelfont={it}, labelsep=period, font=small}
                         \caption{Comparison of hyper-parameter tuning approaches in terms of computational time and the average error obtained using the selected parameter.}
    \vskip -0.15in
     \label{tb:hyper_parameter_tuning}
\setstretch{1.2}
\begin{center}
\scalebox{0.85}{\begin{tabular}{|c|c|c|c|c|}
\hline
\multirow{2}{*}{Data set} & \multicolumn{2}{|c|}{Upper bound approach} & \multicolumn{2}{|c|}{Cross-validation approach} \\
\cline{2-5}
& Error & Time (in secs) & Error & Time (in secs) \\
\hline
Mammographic      		   & 0.21 $\pm$ 0.02	& \textbf{7.6 $\pm$ 0.2}  & 0.21 $\pm$ 0.02  & 69.9 $\pm$ 1.8 \\
Haberman				   & 0.28 $\pm$ 0.04	& \textbf{5.8 $\pm$ 0.7}  & 0.28 $\pm$ 0.05  & 51.4 $\pm$ 2.5 \\
Indian\textunderscore liver	   & 0.29 $\pm$ 0.02  & \textbf{7.5 $\pm$ 0.3} & 0.29 $\pm$ 0.02  & 68.0 $\pm$ 5.0 \\
Diabetes      			   & 0.28 $\pm$ 0.03	& \textbf{8.1 $\pm$ 0.8}  & 0.28 $\pm$ 0.02  & 74.0 $\pm$ 7.4 \\
\hline
\end{tabular}}
\end{center}
 \vskip -0.25in
\end{table}

The results show that both methods obtain similar errors while the approach using upper bound provides significant improvement in the computational time than the usual cross-validation approach. In particular, we observe that the upper bound approach is around 10 times faster than the usual approach. This efficiency is achieved as the 10-fold cross-validation approach requires training 10 \ac{MRC} classifiers to estimate the error corresponding to each parameter value while the upper bound approach requires training a single \ac{MRC} classifier. 

\subsection{Classification with high-dimensional biological data}

\begin{figure}
\centering
 \psfrag{cvx}[l][l][1]{\code{cvx}}
 \psfrag{subgradabcde}[l][l][1]{\code{subgrad}}
 \psfrag{cg}[l][l][1]{\code{cg}}
 \psfrag{0}[][][0.8]{0}
 \psfrag{10}[][][0.8]{10}
 \psfrag{20}[][][0.8]{20}
 \psfrag{30}[][][0.8]{}
  \psfrag{40}[][][0.8]{40}
  \psfrag{50}[][][0.8]{}
  \psfrag{60}[][][0.8]{60}
  \psfrag{70}[][][0.8]{}
  \psfrag{80}[][][0.8]{80}
  \psfrag{1}[][][0.8]{1}
  \psfrag{2}[][][0.8]{2}
  \psfrag{3}[][][0.8]{3}
  \psfrag{4}[][][0.8]{4}
  \psfrag{0.5}[][][0.8]{}
  \psfrag{1.5}[][][0.8]{}
  \psfrag{2.5}[][][0.8]{}
  \psfrag{3.5}[][][0.8]{}
 \psfrag{x}[c][t][1]{Number of features}
 \psfrag{y}[c][t][1]{Time [s]}
\includegraphics[width=0.8\textwidth]{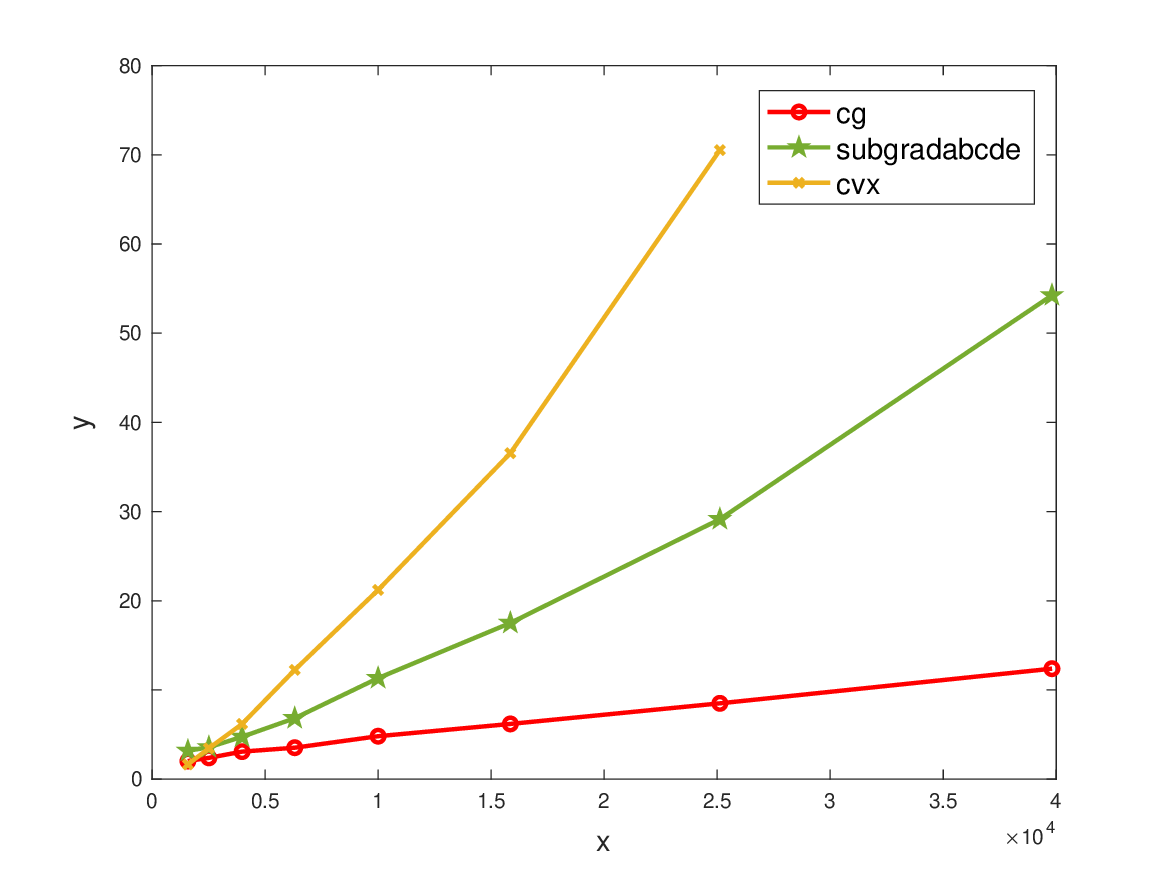}
\captionsetup{font=small}
\caption{Training times using different solvers for "prostate" data set}
\label{fig:solver_times}
\end{figure}

\kar{In the following, we present an example of usage of \pkg{MRCpy} for high-dimensional medical applications. In such applications, the training time of the classification methods is large as the number of parameters of the model increase with the number of dimensions. The example uses the efficient learning algorithm \citep{BonMazPer23} in the library to enable efficiency in these settings. In particular, the following examples illustrates the classification of patients as having prostate cancer or not based on a large number of medical parameters. In addition, we compare the efficiency achieved with other standard solvers such as \code{cvx} and \code{subgrad} (Figure~\ref{fig:solver_times}). Finally, we also show comparison with the related libraries, such as \pkg{mRMR} \citep{DinPen:05} and efficient learning algorithm for \ac{SVM} presented in \cite{DedAntEtal:22}, using multiple biological data sets obtained from \pkg{OpenML} respository (Table~\ref{tb:feature_selection}).}

\kar{Firstly, we load the high-dimensional biological data set ``prostate''. The data set consists of the medical records of 102 patients where each record consists of 12,600 attributes. The data set can be obtained from the \pkg{OpenML} repository using the function} \code{fetch_openml()} \kar{of} \pkg{scikit-learn}:
\begin{Code}
>>> from sklearn.datasets import fetch_openml
>>> X, Y = fetch_openml(name = "prostate", return_X_y = True, 
... version = 1, parser = "auto")
\end{Code}
\kar{We then normalize the data set using the function \pkg{StandardScaler()} of \pkg{scikit-learn} as:}
\begin{Code}
>>> from sklearn.preprocessing import StandardScaler
>>> X = StandardScaler().fit_transform(X)
\end{Code}
\kar{Now, we import the \class{MRC} classifier from \pkg{MRCpy}, and define an instance \code{clf1} corresponding to non-deterministic classification rule using \mbox{0-1} loss and the efficient learning algorithm:}
\begin{Code}
>>> from MRCpy import MRC
>>> clf1 = MRC(loss = "0-1", solver = "cg", deterministic = False,
... random_state = 42)
\end{Code}
In addition, we also define \class{MRC} classifier instances \code{clf2}, \code{clf3} with same attributes as \code{clf1} but using other solvers:
\begin{Code}
>>> clf2 = MRC(loss = "0-1", solver = "cvx", deterministic = False, 
... random_state = 42)
>>> clf3 = MRC(loss = "0-1", solver = "subgrad", max_iters = 80000,
... deterministic = False, random_state = 42)
\end{Code}
\kar{Then, the three classifiers are trained on the ``prostate'' data set and the training times are obtained using \pkg{time} library for comparison:}
\begin{Code}
>>> import time
>>> start_time_clf1 = time.time()
>>> clf1.fit(X,Y)
>>> print("Time taken by cg solver: ", time.time() - start_time_clf1)
>>> start_time_clf2 = time.time()
>>> clf2.fit(X,Y)
>>> print("Time taken by cvx solver: ", time.time() - start_time_clf2)
>>> start_time_clf3 = time.time()
>>> clf3.fit(X,Y)
>>> print("Time taken by subgrad solver: ", time.time() - start_time_clf3)
\end{Code}
\begin{CodeOutput}
Time taken by cg solver: 1.87958683013916
Time taken by cvx solver: 13.698040237426758
Time taken by subgrad solver: 20.975523118972778
\end{CodeOutput}
\kar{Moreover, the worst-case error probability obtained by the different solvers can be compared using the upper bound given by the library:}
\begin{Code}
>>> print("Upper bound using cg solver: ", clf1.get_upper_bound())
>>> print("Upper bound using cvx solver: ", clf2.get_upper_bound())
>>> print("Upper bound using subgrad solver: ", clf3.get_upper_bound())
\end{Code}
\begin{CodeOutput}
Upper bound using cg solver: 0.04791407865441489
Upper bound using cvx solver: 0.04765561349025771
Upper bound using subgrad solver: 0.09436275096765279
\end{CodeOutput}
\kar{The example above shows that the efficient learning algorithm outperforms the other standard solvers in terms of training time with comparable worst-case error probability on the prostate cancer data set.}

Figure~\ref{fig:solver_times} presents the average training time using \code{cg} solver in comparison with other solvers \code{cvx} and \code{subgrad} for 0-1 \class{MRC} classifier in the library. The results are obtained for increasing number of Fourier features for ``prostate'' data set in the range of 100 to 30,000. The results show that for large number of features, for instance 10,000 and more in this data set, the \code{cg} solver can enable efficient learning of \acp{MRC} compared to the other solvers.

\kar{
\begin{table*}
 \captionsetup{labelfont={it}, labelsep=period, font=small}
                         \caption{Comparison of \pkg{MRCpy} with other libraries for high-dimensional application using multiple real biological data sets.}
    \vskip -0.15in
     \label{tb:feature_selection}
\setstretch{1.2}
\begin{center}
\scalebox{0.75}{\begin{tabular}{|c|c|c|c|c|c|c|c|c|c|}
\hline
\multicolumn{1}{|c|}{\multirow{3}{*}{Data set}} & \multicolumn{4}{c|}{\pkg{MRCpy}}& \multicolumn{3}{c|}{\pkg{SVM-CG}} & \multicolumn{2}{c|}{\pkg{mRMR}} \\
 \cline{2-10}
  & {\multirow{2}{*}{$\up{R}^*$}} & {\multirow{2}{*}{Error}} & Training  & No. of & {\multirow{2}{*}{Error}} & Training & No. of & {\multirow{2}{*}{Error}} & Training  \\
  &  &  & time (in secs) & features &  & time (in secs) & features &  & time (in secs) \\
 \hline
Leukemia & .03 & .02 $\pm$ .05  & 1.0 $\pm$ 0.0 & \ 64 $\pm$ 0 & .07 $\pm$ .11 & 0.2 $\pm$ 0.0 & 37 $\pm$ 2 & .02 $\pm$ .05 & 23.5 $\pm$ 2.4 \\

Ovarian & .03 & .00 $\pm$ .00  &  5.3 $\pm$ 1.2 & 187 $\pm$ 9 & .00 $\pm$ .00 & 2.4 $\pm$ 0.0 & 33 $\pm$ 2 & .00 $\pm$ .00 & 49.6 $\pm$ 3.3\\

Colon 	& .06 & .11 $\pm$ .13 & 0.5 $\pm$ 0.0 & \ 55 $\pm$ 0 & .18 $\pm$ .09 & 0.2 $\pm$ 0.0 & 33 $\pm$ 2 & .22 $\pm$ .12 & \ 8.1 $\pm$ 1.5\\

Prostate & .04 & .05 $\pm$ .06 &  1.6 $\pm$ 0.1 & \ 91 $\pm$ 0 & .07 $\pm$ .11 & 1.1 $\pm$ 0.1 & 48 $\pm$ 2 & .06 $\pm$ .06 & 80.8 $\pm$ 4.0\\
\hline
\end{tabular}}
\end{center}
\vskip -0.25in
\end{table*}

Table~\ref{tb:feature_selection} presents the comparison of \pkg{MRCpy} with other libraries for high-dimensional medical applications. In particular, we compare with the \pkg{mRMR} library \citep{DinPen:05} for feature selection, and the efficient learning approach presented in \cite{DedAntEtal:22} for \ac{SVM} (SVM-CG) in terms of 10-fold cross-validated classification error, training time, and number of features selected. The comparison is using the real world biological data sets obtained from the \pkg{OpenML} repository. Note that the results for \pkg{mRMR} library (for feature selection) are obtained using the \ac{SVM} classifier and for the same number of features as \pkg{SVM-CG} library.

The results show that \pkg{MRCpy} can enable efficient learning in high-dimensions compared to the feature selection library \pkg{mRMR} and is competitive with \pkg{SVM-CG}. Moreover, \pkg{MRCpy} can provide alternate performance assessment in terms the worst-case error probability $\up{R}^{*}$ without cross-validation. Note that the cross-validation error can be reliable in these applications due to large standard deviations as shown in Table~\ref{tb:feature_selection}.
}

\subsection{Classification under concept drift}
\label{sec:results_concept_drift}
In the following, we present an example to show the usage of \class{AMRC} class in \pkg{MRCpy} for supervised classification under concept drift. 

Methods implemented using class \class{AMRC} are evaluated using ``Usenet2'' data set that has been often used as benchmarks for supervised classification under concept drift \citep{zhao2020handling}. ``Usenet2'' is a real data set which consists of 1,500 instances with 100 attributes based on 20 newsgroups collection. The data set simulates a stream of messages from different newsgroups that are sequentially presented to a user and the goal is to predict the personal interests. Firstly, we load the ``Usenet2'' data set using the function \code{load\_usenet2()}, which returns 2 arrays that correspond to instances and labels. Each row in the first array contains an instance, while each column contains an attribute.
\begin{Code}
>>> from MRCpy.datasets import load_usenet2
>>> X, Y = load_usenet2()
\end{Code}
Now, we import the \class{AMRC} class from \pkg{MRCpy} library. Then, we define an instance \code{clf} for binary classification using the Fourier feature mapping, 0-1 loss, and randomized classification rule.
\begin{Code}
>>> from MRCpy import AMRC
>>> clf = AMRC(n_classes = 2, phi = "fourier", random_state = 42)
\end{Code}
At each time, we update the classification rule using the most recent instance-label pair. Then, we predict a label, obtain the upper bound for error probability, and obtain the bound for accumulated mistakes as follows.
\begin{Code}
>>> n = X.shape[0]
>>> import numpy as np
>>> bound_error_probability = np.zeros(n - 1)
>>> bound_accumulated_mistakes = np.zeros(n - 1)
>>> Y_pred = np.zeros(n - 1)
>>> error = np.zeros(n - 1)
>>> for i in range(n - 1):
...    clf.fit(X[i, :], Y[i])
...    bound_error_probability[i] = clf.get_upper_bound()
...    bound_accumulated_mistakes[i] = clf.get_upper_bound_accumulated()
...    Y_pred[i] = clf.predict(X[i + 1, :])
...    error[i] = (Y[i+1] != Y_pred[i])
>>> print("Classification error: ", np.mean(error))
\end{Code}
\begin{CodeOutput}
Classification error: 0.3055370246831221
\end{CodeOutput}

\begin{figure}
\centering
 \psfrag{Bound AMRC}[l][l][1]{Bound AMRC $\delta = 0.05$}
 \psfrag{AMRC}[l][l][1]{AMRC}
 \psfrag{Deterministic AMRCabcdefg}[l][l][1]{Deterministic AMRC}
 \psfrag{Accumulated mistakes per time}[][][1]{Accumulated mistakes per time}
 \psfrag{Time}[][][1]{Time $t$}
 \psfrag{0}[][][0.8]{0}
 \psfrag{500}[][][0.8]{500}
 \psfrag{1000}[][][0.8]{1000}
 \psfrag{1500}[][][0.8]{1500}
  \psfrag{0.25}[][][0.8]{0.25}
  \psfrag{0.5}[][][0.8]{0.5}
  \psfrag{0.75}[][][0.8]{0.75}
  \psfrag{1}[][][0.8]{1}
\includegraphics[width=0.8\textwidth]{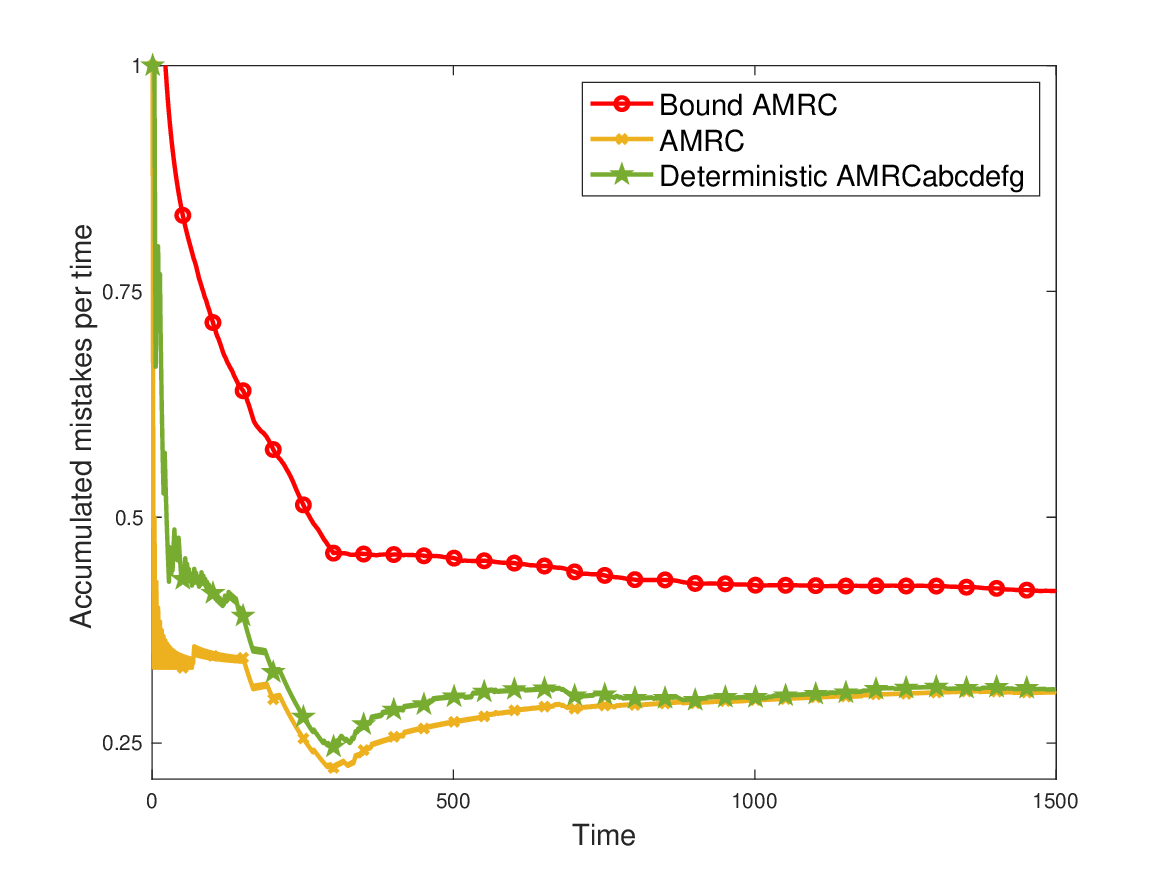}
\captionsetup{font=small}
\caption{Results on ``Usenet2'' data set shows the evolution of accumulated
mistake bounds and accumulated mistakes per number
of steps.}
\label{fig:bound_AMRC}
\end{figure}

\begin{table}
 \captionsetup{labelfont={it}, labelsep=period, font=small}
 \renewcommand\thetable{5}
                         \caption{Accuracies in [\%] obtained by \pkg{MRCpy} and \pkg{River} libraries for supervised classification under concept drift using multiple real data sets. }
     \label{tb:amrc}
     \vskip -2in
\setstretch{1.2}
\begin{center}
\scalebox{1}{\begin{tabular}{|l|c|c|c|}
\hline
\multirow{ 2}{*}{Data set}                           & \pkg{River} & \pkg{MRCpy} & \pkg{MRCpy} \\
& & (Deterministic \class{AMRC}) & (\class{AMRC}) \\
\hline
Agrawal-abrupt  & 52.00 & 67.68 & 68.69\\
Airlines   & 74.00 & 77.78 & 78.79 \\
SMTP & 98.00 & 100.00 & 100.00 \\
HTTP & 99.00 & 100.00 & 100.00 \\
Credit & 71.00 & 76.77 & 74.75 \\
Usenet2 & 63.00 & 56.57 & 67.68 \\
\hline
\end{tabular}}
\end{center}
\end{table}

The example above shows that the classification error is 0.36 using Fourier feature mapping and randomized classifier (see details in~\cite{AlvMazLoz22}). Figure~\ref{fig:bound_AMRC} shows the accumulated mistakes per time step of \ac{AMRC} method in comparison with the bounds for accumulated mistakes per time. Such figure shows that \ac{AMRC} method provides tight performance guarantees in terms of bounds for accumulated mistakes. 

Table~\ref{tb:amrc} shows the accuracy of the proposed \ac{AMRC} method in comparison with the method proposed in \pkg{River} library~\citep{montiel2021river}. These methods are evaluated using multiple data sets that have been often used as benchmarks for supervised classification under concept drift \citep{kolter2007dynamic, gomes2017adaptive, nguyen2018variational}: ``Agrawal-abrupt'', ``Airlines'', ``SMTP'', ``HTTP'', and ``Credit card''. The benchmark data sets can be obtained from \pkg{River}~\citep{montiel2021river} and \pkg{scikit-multiflow}~\citep{skmultiflow} libraries. The numerical results in Table~\ref{tb:amrc} are obtained using the first hundred instances of each data set. Such table shows that deterministic and non-deterministic \acp{AMRC} offer an overall improved performance compared to the method in \pkg{River} library along the benchmark data sets.

\subsection{Classification under general covariate shift}\label{Sec:Examples_DWGCS}
In the following, we present an example to show the usage of \class{DWGCS} class in \pkg{MRCpy} for covariate shift adaptation in supervised classification.

\jose{Methods implemented using class \class{DWGCS} are evaluated using ``20 Newsgroups'' data set, available at \url{http://qwone.com/~jason/20Newsgroups/}.
This data set is intrinsically affected by a covariate shift since the training and testing partitions correspond to different times, and it has been often used as a benchmark for covariate shift adaptation \citep{Zhang2013,Sakai2019,Sun2011}.
The data set is a collection of around 20,000 newsgroups documents, partitioned across 20 different categories. Some of the categories are closely related to each other so that we group the categories in four different classes: comp, sci, talk and rec.
For these experiments, we consider 4 binary classification problems, as in \citep{Zhang2013}, utilize 500 features with highest Pearson's correlation, and randomly sample 500 training samples and testing instances in each repetition.

Existing reweighted and robust approaches require very strong assumptions regarding the support of the training and testing distributions.
Reweighted techniques require that the support of the training distribution contains that of testing, and robust techniques require that the support of the testing distribution contains that of training.
In the proposed experiments, as it happens in practice, the distribution of the training and testing instances differs in an arbitrary manner, so that the supports may not be contained in each other.

Firstly, we load the ``comp vs sci short'' data set for covariate shift adaptation on supervised classification using \code{load\_comp\_vs\_sci\_short()} function, that returns two arrays composed by instances and labels from the training distribution and two arrays composed by instances and labels from the testing distribution.}
\begin{Code}
>>> from MRCpy.datasets import load_comp_vs_sci_short
>>> X_train, Y_train, X_test, Y_test = load_comp_vs_sci_short()
\end{Code}
Now, we import the \class{DWGCS} class from \pkg{MRCpy}, and define an instance \code{clf}:
\begin{Code}
>>> from MRCpy import DWGCS
>>> clf = DWGCS(loss = "0-1", phi = "linear", deterministic = True) 
\end{Code}
Then, we compute the weight functions $\alpha(x)$ and $\beta(x)$ (discussed in Section~\ref{Sec:DWGCS}), and learn the \ac{DW-GCS} classification rule using the \code{fit()} function. The \code{fit()} function takes as input the training instance-label pairs \code{X_train} and \code{Y_train}, along with the testing instances \code{X_test} available at learning in covariate shift scenarios.
These testing instances are used in order to compute weights $\alpha(x)$ and $\beta(x)$, and to solve the minimax risk problem.
\begin{Code}
>>> clf.fit(X_train, Y_train, X_test)
>>> Y_pred = clf.predict(X_test)
>>> print("Classification error using 0-1 loss: ", clf.error(X_test, Y_test))
>>> print("Minimax risk: ", clf.get_upper_bound())
\end{Code}
\begin{CodeOutput}
Classification error using 0-1 loss: 0.223
Minimax risk: 0.08770133070040598
\end{CodeOutput}
\jose{Figure~\ref{fig:DWGCS} shows box-plots corresponding to the classification error obtained by the \ac{DW-GCS} method implemented in the \class{DWGCS} of the \pkg{MRCpy} library compared to that obtained without covariate shift adaptation. The results are obtained for 4 different classification tasks using \mbox{``20 Newsgroups''} data set. The figure compares the \ac{DW-GCS} method with \acp{MRC} implemented in class \class{CMRC} that does not take into account covariate shift, the \class{DWGCS} class with \code{D = 1} that corresponds to the reweighted approach for covariate shift, and with \mbox{\code{D = np.inf}} that corresponds to the robust approach for covariate shift discussed in Section~\ref{Sec:DWGCS}. For the case where \mbox{\code{D = np.inf}}, in order to obtain weights $\alpha(x)$ corresponding to the robust approach, we solve the conventional \ac{KMM} problem \citep{HuaSmoGre:06}, assigning weights to the testing instances instead of the training instances, and restricting the weights $\alpha(x)$ in the optimization problem to be less or equal than 1.

\begin{figure}[ht]
\centering
 \psfrag{Experiment}[][][1]{}
 \psfrag{Classification error}[][][1]{Classification error}
 \psfrag{0.1}[][][0.8]{0.1}
 \psfrag{0.15}[][][0.8]{0.15}
 \psfrag{0.2}[][][0.8]{0.2 }
 \psfrag{0.25}[][][0.8]{0.25}
 \psfrag{0.3}[][][0.8]{0.3 }
 \psfrag{2}[][][0.8]{comp vs sci}
 \psfrag{4}[][][0.8]{comp vs talk}
 \psfrag{6}[][][0.8]{rec vs talk}
 \psfrag{8}[][][0.8]{sci vs talk}
 \psfrag{No adapt.}[l][l][0.8]{\hspace{0.1cm}CMRC}
 \psfrag{titulo grande}[l][l][0.8]{\hspace{0.1cm}KMM}
 \psfrag{titulo}[l][l][0.8]{\hspace{0.1cm}Robust}
 \psfrag{DW-GCS}[l][l][0.8]{\hspace{0.1cm}DWGCS}
\includegraphics[width=0.78\textwidth]{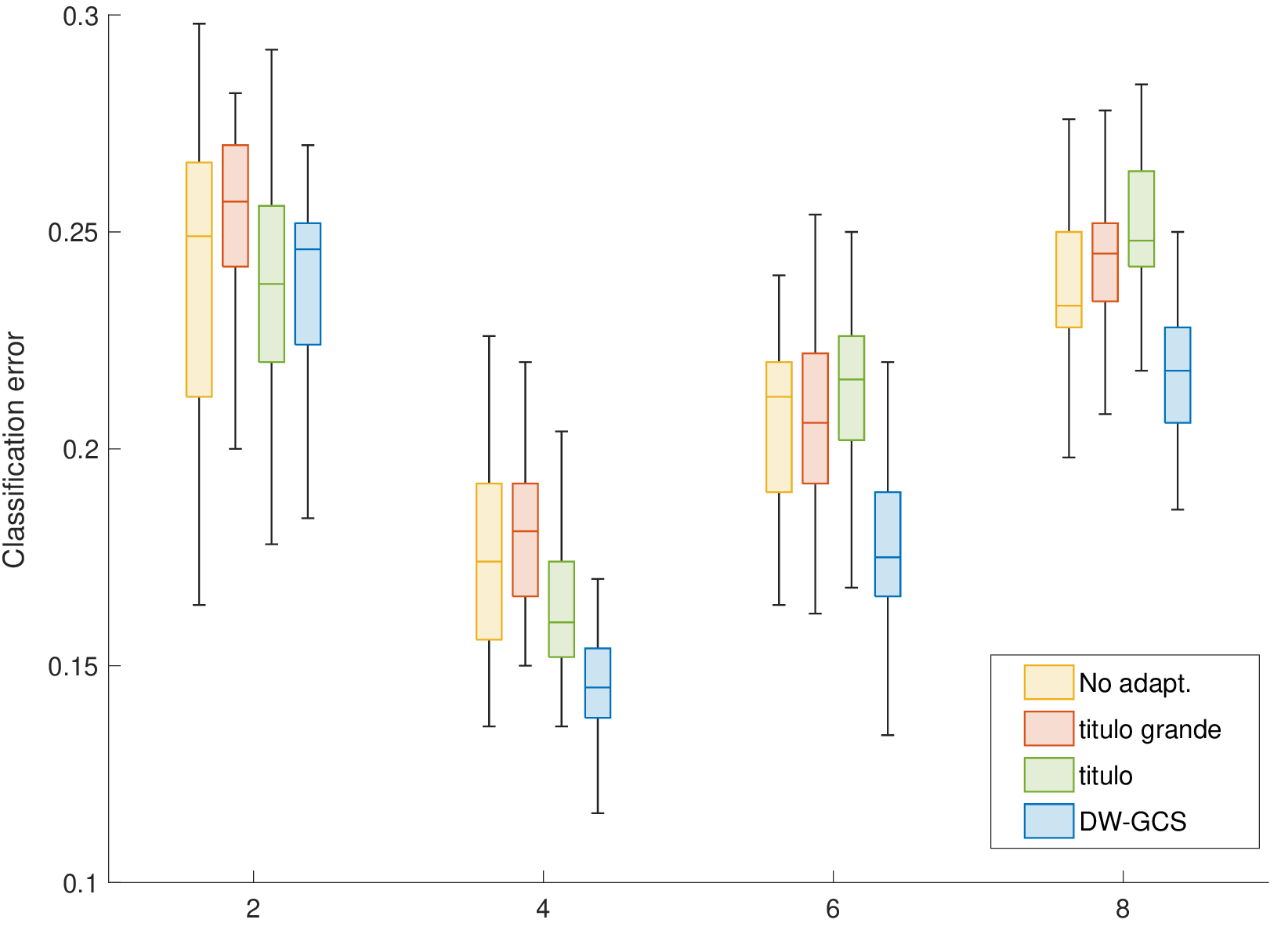}
\captionsetup{font=small}
\caption{Results on 4 different binary classification tasks using ``20 Newsgroups'' data set shows that the DW-GCS method implemented in the library can adapt to general covariate shift.}
\label{fig:DWGCS}
\end{figure}


The results show that the \ac{DW-GCS} method implemented in the library can more adequately adapt to general covariate shift. In particular, Figure~\ref{fig:DWGCS} shows that solving the \ac{DW-KMM} problem instead of using the existing \ac{KMM} problem (that fix weights $\alpha(x)$ to be equal 1) results in significant performance improvement when the support of the distributions are not contained in each other.}

\section{Closing remarks} \label{sec:concl}
We presented a \proglang{Python} library \pkg{MRCpy} for recently proposed \acp{MRC} based on \ac{RRM} approach which can provide performance guarantees and easily adapt to distribution shifts. The presented library implements multiple variants of \acp{MRC} that can enable efficient learning in high dimensions and adapt to distribution shifts such as covariate shift and concept drift. \pkg{MRCpy} is designed with an object-oriented approach that facilitates adaptability and the code follows the standards of popular machine learning library that facilitates readability and easy usage. The presented library is available under the GPL-3.0 license on GitHub at \url{https://github.com/MachineLearningBCAM/MRCpy}. The library undergoes continuous testing with upcoming updates and is actively maintained.

\section*{Acknowledgements}
Funding in direct support of this work has been provided by projects PID2022-137063NB-
I00 and CNS2022-135203 funded by MCIN/AEI/10.13039/501100011033 and the European Union “NextGenerationEU”/PRTR, BCAM Severo Ochoa accreditation CEX2021-001142-S / MICIN / AEI/ 10.13039/501100011033 funded by the Ministry of Science and Innovation, and programes ELKARTEK and
BERC-2022-2025 funded by the Basque Government. Kartheek Bondugula also holds a predoctoral grant (EJ-GV 2022) from the Basque Government.



\newpage

\end{document}